\documentclass[10pt,twocolumn,letterpaper]{article}

\PassOptionsToPackage{table,xcdraw}{xcolor} 
\usepackage{xcolor}

\usepackage{bm}
\usepackage{iccv}
\usepackage{times}
\usepackage{booktabs}
\usepackage{graphicx}
\usepackage{amsmath}
\usepackage{amssymb}
\usepackage{tikz}
\usetikzlibrary{calc}
\usetikzlibrary{positioning}
\usetikzlibrary{shapes.geometric, arrows, arrows.meta}
\tikzstyle{arrow} = [thick,->,>=stealth]
\tikzset{>={Stealth[width=5mm]}}
\tikzstyle{point} = [circle, text centered, draw=black, align=center, font=\small]
\usepackage{multirow}
\definecolor{iccvblue}{rgb}{0.21,0.49,0.74}
\usepackage{multicol}
\usepackage[ruled,linesnumbered]{algorithm2e}
\usepackage{indentfirst}
\usepackage{bbding}
\usepackage{pifont}
\usepackage{arydshln}
\usepackage{diagbox}

\usepackage{tikz}
\usetikzlibrary{calc}
\usetikzlibrary{positioning}
\usetikzlibrary{shapes.geometric, arrows, arrows.meta}
\tikzstyle{arrow} = [thick,->,>=stealth]
\tikzset{>={Stealth[width=5mm]}}
\tikzstyle{point} = [circle, text centered, draw=black, align=center, font=\small]
\usepackage[ruled,linesnumbered]{algorithm2e}

\usepackage[pagebackref,breaklinks,colorlinks]{hyperref}

\usepackage[capitalize]{cleveref}
\crefname{section}{Sec.}{Secs.}
\Crefname{section}{Section}{Sections}
\Crefname{table}{Table}{Tables}
\crefname{table}{Tab.}{Tabs.}

 \iccvfinalcopy 
\def\iccvPaperID{4896} 

\ificcvfinal\fi

\begin{document}

\title{Human-Imperceptible Physical Adversarial Attack for NIR Face Recognition Models}

\author{
Songyan Xie$^{1}$, \quad Jinghang Wen$^{2}$, \quad Encheng Su$^{3}$, \quad Qiucheng Yu$^{2}$\thanks{Corresponding author} \\
$^{1}$ School of Computer Science and Technology, China University of Mining and Technology, Xuzhou, China\\
$^{2}$ Department of Computer Science, City University of Hong Kong, Hong Kong, China\\
$^{3}$ Department of Engineering and Design, Technische Universität München, München, Germany\\
\tt\small xiesongyan@cumt.edu.cn, jh.wen@my.cityu.edu.hk, encgo.su@tum.de, qiuchenyu2-c@my.cityu.edu.hk
}

\maketitle
\ificcvfinal\thispagestyle{empty}\fi

\begin{abstract}
    Near-infrared (NIR) face recognition systems, which can operate effectively in low-light conditions or in the presence of makeup, exhibit vulnerabilities when subjected to physical adversarial attacks. To further demonstrate the potential risks in real-world applications, we design a novel, stealthy, and practical adversarial patch to attack NIR face recognition systems in a black-box setting. We achieved this by utilizing human-imperceptible infrared-absorbing ink to generate multiple patches with digitally optimized shapes and positions for infrared images. To address the optimization mismatch between digital and real-world NIR imaging, we develop a light reflection model for human skin to minimize pixel-level discrepancies by simulating NIR light reflection.
 Compared to state-of-the-art (SOTA) physical attacks on NIR face recognition systems, the experimental results show that our method improves the attack success rate in both digital and physical domains, particularly maintaining effectiveness across various face postures. Notably, the proposed approach outperforms SOTA methods, achieving an average attack success rate of 82.46\% in the physical domain across different models, compared to 64.18\% for existing methods.
The artifact is available at \href{https://anonymous.4open.science/r/Human-imperceptible-adversarial-patch-0703/}{https://anonymous.4open.science/r/Human-imperceptible-adversarial-patch-0703/}.
\end{abstract}

\section{Introduction}
Face recognition models have been applied in various tasks, including surveillance~\cite{wang2017face}, border security~\cite{carlos2018facial}, and access control~\cite{apple2023,microsoft2023}. However, traditional visible (VIS) face recognition systems encounter performance degradation under challenging conditions such as dark environments or makeup. To overcome these limitations, near-infrared (NIR) face recognition systems have emerged as a powerful alternative, as they can capture facial features in low-light or complex environments by utilizing the unique properties of infrared light, which is immune to visible lighting and cosmetic alterations~\cite{kong2005recent,krishnan2022facial}.

\begin{figure}[t]
\centering
\input{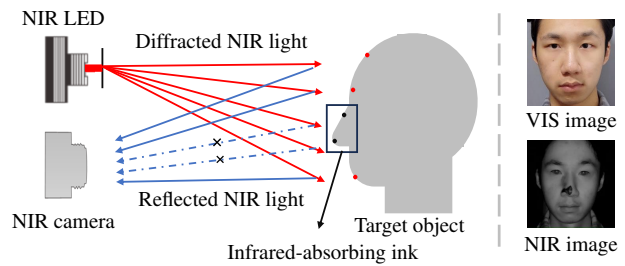}
\caption{The diagram illustrates the theory of the NIR face recognition system and our proposed attack strategy. The left subfigure shows that infrared-absorbing ink can absorb diffracted NIR light and reduce the reflected NIR light used for imaging. The right subfigure presents our adversarial ink patch. The VIS image, perturbed with the ink, remains imperceptible, while the NIR image successfully executes an adversarial attack.}
\label{fig: NIR face recognition systems attack}
\end{figure}

Modern NIR face recognition systems leverage deep neural networks (DNNs) to extract illumination-robust features and establish cross-modal mappings between NIR and VIS images~\cite{fu2021dvg,wu2018lightCNN}. While these DNN-driven architectures achieve state-of-the-art recognition accuracy, their inherent vulnerability to adversarial perturbations~\cite{szegedy2013intriguing} raises critical security concerns, since a small perturbation added to the input can lead to incorrect identity judgments. Previous work has successfully achieved physical adversarial attacks on VIS face recognition systems by embedding adversarial perturbations into common accessories, such as hats~\cite{sharif2016accessorize}, glasses~\cite{sharif2016accessorize}, and masks~\cite{zolfi2022adversarial}, which mislead the system into predicting the wrong identity. Considering the concealment requirement of adversarial patches, some other works exploit imperceptible optical interference mechanisms, including infrared light illumination manipulation~\cite{WangTheIP} and persistence-of-vision effects~\cite{shen2019vla}, which bypass human visual detection while compromising VIS face recognition systems. To explore the vulnerabilities of NIR face recognition systems,  research~\cite{cohen2023accessorize} reveals that printable adversarial eyeglasses can effectively deceive NIR face recognition systems by minimizing color derivation between digital simulations and physical deployments in the NIR spectrum.

All the aforementioned works primarily focus on VIS face recognition adversarial attacks. Although research~\cite {cohen2023accessorize} has achieved successful attacks on NIR face recognition systems, it overlooked the stealthiness and real-world feasibility. These limitations stem from three fundamental challenges inherent to NIR face recognition systems attacks. (i) Most previous studies targeting VIS face recognition attacks cannot mislead NIR face recognition systems, as NIR images are robust to lighting-variation due to specialized wavelength filters and exhibit lower dimensionality correlating with reduced susceptibility to attacks~\cite{shamir2019simpleexplanationexistenceadversarial}. (ii) Some other works generated perturbations in NIR images, but the strange patch patterns appeared significantly suspicious in the VIS environment, which fails to meet the concealment needs for attacks. (iii) The only study~\cite{cohen2023accessorize} that conducted a white-box attack on the NIR face recognition model, which requires complete knowledge of the model's structure and parameters, conflicts with real-world deployments black-box constraints where only the model's predicted label and probability are accessible.

To address the above challenges, we propose a novel human-imperceptible physical adversarial patch designed to deceive NIR face recognition systems. As illustrated in \cref{fig: NIR face recognition systems attack}, our method employs transparent infrared-absorbing ink to generate a human-imperceptible adversarial patch that effectively disrupts NIR facial imaging. Furthermore, we introduce a comprehensive attack framework that jointly optimizes the shapes and positions of patches under strictly black-box settings. By simulating NIR light reflection during optimization, our patch is refined to achieve the physical optimal shape and position, resulting in an average attack success rate of 82.46\% across different models in the physical domain, compared to 64.18\% for the existing state-of-the-art method.

Our main contributions are summarized as follows.

\begin{itemize}
    \item To achieve a stealthy and effective adversarial attack for NIR face recognition systems, we propose the first human-imperceptible physical adversarial patch utilizing infrared-absorbing ink. This method ensures the patch is imperceptible in the VIS spectrum while effectively misleading NIR face recognition models.

    \item To optimize the patch under strict black-box conditions, we develop a multi-patch adversarial attack framework. This framework jointly optimizes patch shape and position relying on model prediction labels and probabilities, eliminating the need for white-box access and enhancing real-world applicability.

    \item To bridge the gap between digital simulation and real-world NIR imaging, we introduce a light reflection model that simulates NIR light reflection with human skin. This model guides the optimization process toward the physically optimal patch configuration, ensuring its effectiveness after physical deployment.
\end{itemize}

\begin{figure*}[t]
\centering
\input{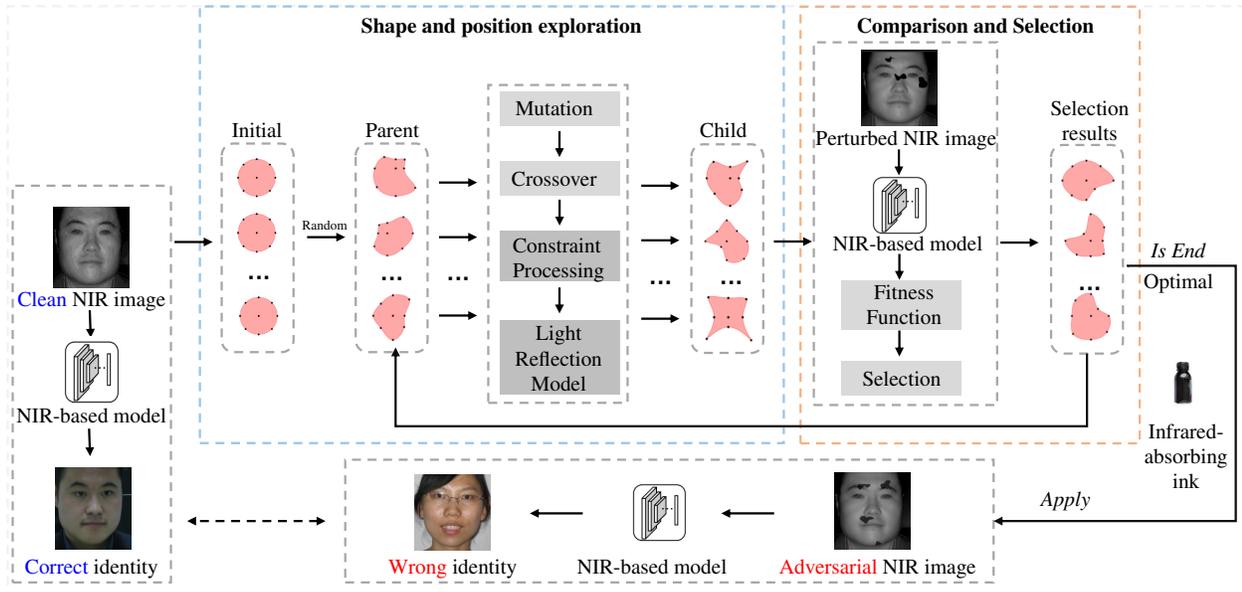}

\caption{An overview of generating NIR face recognition adversarial patches using the Differential Evolution (DE) framework: The process starts with an initial population of shapes, which undergo shape and position exploration to generate a child population. A fitness function evaluates both parent and child populations, selecting the best individuals to enhance adversarial effectiveness. Finally, the optimal individual is then applied to the NIR image for the attack.}

\label{fig: attack framework}
\end{figure*}

\section{Background and Related Work}
\label{sec:formatting}
In this section, we first provide an overview of NIR face recognition systems, followed by a review of existing adversarial attacks in both the digital and physical domains.

\subsection{Face Recognition in NIR}
\label{sec: NIR background}
Near-infrared (NIR) refers to light in the electromagnetic spectrum, with wavelengths ranging from 750 nm to 1400 nm. While existing research predominantly focuses on the visible spectrum face recognition systems, the integration of NIR imaging significantly enhances robustness against pose variations, illumination changes, and partial occlusions, owing to NIR's inherent insensitivity to visible lighting fluctuations and cosmetic alterations~\cite{hu2017heterogeneous}.  Additionally, the reflection difference between live spoof faces under NIR can be exploited to detect presentation attacks such as print, video replay, makeup, and 3D mask \cite{jiang2019multilevel}. These advantages have led to the security-critical adoption of NIR face recognition systems \eg, \cite{apple2023,microsoft2023}.

Leading NIR face recognition systems rely on DNNs\cite{fu2019dual, yu2021lamp, fu2021dvg, hu2021dual, miao2022physically, wu2018lightCNN}. For example, Wu \etal proposed the LightCNN framework, designed for deep face representation using large-scale datasets with noisy labels, which was optimized for efficiency and adaptability across various face recognition tasks \cite{wu2018lightCNN}. Based on this, Fu \etal introduced LightCNN-DVG \cite{fu2019dual}, which achieved state-of-the-art performance in NIR face recognition. The key innovation of LightCNN-DVG lies in its fine-tuning process with NIR-VIS data, ensuring identity consistency through distribution alignment in the latent space and pairwise identity preservation in the feature space while mapping features of different identities further apart. However, despite their high accuracy in face recognition tasks, these NIR-based models remain susceptible to adversarial attacks. 

\subsection{Digital Adversarial Attacks}
\label{sec: digital adversarial attacks}
Adversarial attacks can be divided into white-box attacks and black-box attacks. In the white-box setting, where attackers have complete knowledge of the model's structure and parameters, attacks are performed using optimization techniques, such as Box-constrained L-BFGS \cite{szegedy2013intriguing}, C\&W \cite{carlini2017towards}, \etc. PGD \cite{madry2017towards} is a multi-step iterative method that uses projected gradient descent on the negative loss function to generate adversarial examples. In contrast, black-box attacks do not require detailed model parameters and can be divided into three categories: transfer-based, score-based, and decision-based. For transfer-based methods, adversarial examples generated for one model can be transferred to another model, leading to successful attacks \cite{dong2018boosting,liu2016delving,chen2020universal}. Score-based methods, where the target model’s predicted labels and probabilities are accessible, often employ gradient estimation \cite{chen2017zoo} and random search \cite{croce2022sparse}. Decision-based methods are more suitable for restrictive scenarios since only the model decisions are available \cite{tao2023hard}. Dong \etal \cite{dong2019efficient} perform digital attacks on face recognition systems in this setting, modeling the local geometry of solution directions to improve efficiency. Attacks aimed at changing the predicted class from the true label are referred to as untargeted attacks (dodging), while attacks targeting a specific class are known as targeted attacks (impersonation). Although digital attacks expose model vulnerabilities, they often fail in real-world scenarios due to physical constraints.

\subsection{Physical Adversarial Attacks}
\label{sec: physical adversarial attacks}
Existing physical attacks on VIS face recognition are divided into printed and light methods. Printed-based adversarial attacks embed perturbation patterns within fixed regions of commonplace accessories such as hats~\cite{komkov2021advhat}, glasses~\cite{sharif2016accessorize}, and masks~\cite{zolfi2022adversarial}, where adversarial optimization explicitly accounts for printer gamut constraints. This compensates for color deviations between digital and physical reproductions, ensuring the realizability of attacks under real-world imaging conditions. However, the printed method inevitably generates visually conspicuous artifacts in adversarial patterns due to unconstrained color-space optimization. To achieve stealthy adversarial attacks, some studies utilize human-imperceptible optical phenomena (\eg, invisible infrared light~\cite{WangTheIP,zhou2018invisible}, persistence of vision~\cite{shen2019vla}) to interfere with captured RGB face images and successfully mislead the VIS face recognition model's output. Although substantial research has focused on adversarial attacks against VIS face recognition systems, these methods have limited transferability to NIR systems due to NIR spectral filtering and lower dimensionality. To address this, research~\cite{cohen2023accessorize} minimizes NIR spectral color discrepancies between digital simulations and physical prints by employing color-constrained adversarial perturbation optimization, thus tackling the mismatch between digital and physical optimality. However, while this work proposes physical printable glasses attacks on NIR face recognition systems, it also introduces conspicuous artifacts in the VIS setting. To the best of our knowledge, no prior work has proposed human-imperceptible physical adversarial attacks for NIR face recognition systems.

\begin{figure*}[t]
\centering
\input{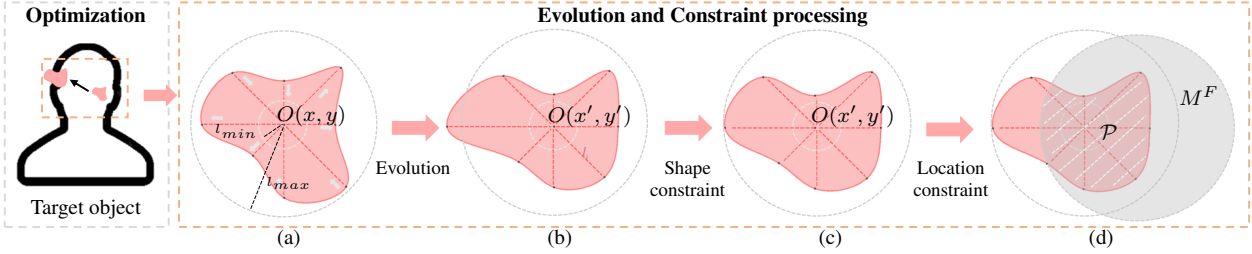}

\caption{The shape and position optimization with constraint process is illustrated as follows: Subfigure (a) depicts the evolution of the shape and its position. Subfigure (b) shows the updated shape after the adjustment in (a), which exceeds the parameter boundaries. Subfigure (c) presents the corrected shape derived from (b). Finally, Subfigure (d) highlights the valid area, denoted as $\mathcal{P}$, constrained within $M^F$.}
\label{fig:shape}

\end{figure*}

\section{Methodology}
In this section, we first briefly introduce the target of adversarial attacks in face recognition, then describe our attack method for optimizing patches in the digital and physical domains.

\subsection{Problem Formulation}
\label{sec: problem formulation}
In the NIR face recognition task, given a clean NIR image $x^{nir}$, the goal of the adversarial attack is to make the NIR face recognition model predict an incorrect identity of the perturbed image $x_{adv}^{nir}$. The adversarial multi-patch perturbation image can be generated using  \cref{eq: gen_adv_image}.
\begin{equation}
\label{eq: gen_adv_image}
x_{adv}^{nir} = (1-\mathcal{M})\odot x^{nir}+\mathcal{M} \odot \hat{x}^{nir}
\end{equation}
where $\odot$ is the Hadamard product, $\hat{x}^{nir}$ is the adversarial multi-patch, and $\mathcal{M} \in \{0, 1\}^{h \times w}$ is a mask matrix to constrain the shape and position of the adversarial patch applied to the target object.



Particularly, there are two types of adversarial attacks on face recognition: dodging (untargeted) attacks and impersonation (targeted) attacks. Given the NIR face recognition model $f(\cdot,\cdot)$, ground-truth label $\hat{t}$ for $x^{nir}$ and $x^{nir}_{adv}$, the goal of a \textbf{dodging} attack is to find the optimal attack parameters $\mathcal{P}_s^*$ to make the probability corresponding to the gallery VIS image $x_{\hat{t}}^{vis}$ as small as possible so that any person different from $\hat{t}$ is regarded as the top-1 identity. So the objective function of the dodging attack can be formalized as:
\begin{equation}\label{eq:untarget}
\min _{\mathcal{P}_s} \ \ \mathcal{L}_{\emph{dodging}}(\mathcal{P}_s)=f(x_{adv}^{nir},x_{\hat{t}}^{vis})
\end{equation}

For the \textbf{impersonation} attack, given a target identity $t^*$, the objective function is defined as follows, which aims to make the probability of the target label as the top-1 identity:
\begin{equation}\label{eq:target}
\min _{\mathcal{P}_s} \ \ \mathcal{L}_{\emph{impersonation}}(\mathcal{P}_s)=1-f(x_{adv}^{nir},x_{t^*}^{vis})
\end{equation}

\subsection{Ink Shaping Attack in the Digital Domain}
\label{sec:Ink shaping attack}
Previous methods have either focused on optimizing the content of adversarial patches~\cite{liu2018dpatch}
 or their positions~\cite{smallbulbs,wei2023hotcold}
, both of which suffer from limited stealthiness and poor effectiveness. Therefore, in this paper, we use location and shape to model a patch and optimize both simultaneously, which significantly enhances the performance of the attack.


\textbf{Patch Location:} To improve the effectiveness of the attack, we employ a multi-patch joint attack strategy. Specifically, we first propose identifying a set of coordinates $\mathcal{O}_s = \{(x_1, y_1), (x_2, y_2), \dots, (x_m, y_m)\}$ to control the central positions of patches $\mathcal{P}_s = \{\mathcal{P}_1, \mathcal{P}_2, \dots, \mathcal{P}_m\}$, where $m$ is the number of patches. This can be formulated as follows:
\begin{equation}
\mathcal{O}_s = \{(x_i, y_i) \mid x_l \leq x_i \leq x_r, y_d \leq y_i \leq y_u| 0 \leq i < m\}
\end{equation}
where $x_l$, $x_r$, $y_d$, and $y_u$ denote the minimum and maximum bounds of the image's width and height. 

Additionally, to determine the valid face mask $M^F\in R^{h\times w}$ for applying patches (\eg, cheeks and forehead, \etc, while excluding critical facial features like the eyes and mouth), we first extract the key facial feature points and then fill the effective region to generate the mask. Therefore, the final patch application areas $\mathcal{M}$ are defined by $\mathcal{P}_i\cap M^F$ for $i=1,2,\dots,m$.


\textbf{Patch Shape.} Most prior studies use a set of vertices to define a polygon~\cite{wei2023unified, zhong2022shadows}. However, this approach will result in overlapping patch contours or excessively large patches. To address these issues, we propose a novel shape representation that implements radial distances and angles of each vertex. Given the number of vertex $n$ and sets of radial distances $\mathcal{L}_s=\{(l_{11},l_{12},\dots, l_{1n})(l_{21},l_{22},\dots, l_{2n})\dots (l_{m1},l_{m2},\dots, l_{mn})\}$, the coordinate $(x_{p_{ij}},y_{p_{ij}})$ of each $j$-th vertex for the $i$-th patch can formulated by:
\begin{equation}
\label{caculate vertexs}
p_{ij}= \begin{cases}
x_{p_{ij}}=x_i+l_{ij}\cdot cos(j\cdot\theta) \\
y_{p_{ij}}=y_i+l_{ij}\cdot sin(j\cdot\theta)\\
l_{\min} \leq l_{ij} \leq l_{\max} 
\end{cases} 
\end{equation}
where $x_i,y_i$ represent the $i$-th patch's central position's coordinates, $\theta=\frac{2\pi}{n}$ indicates the angular spacing between $p_{ij}$ and $p_{ij+1}$, and $l_{min}$ and $l_{max}$ impose constraints on the magnitude of deformations applied to the patch, making the patch appear as if the distortions were caused by camera capture errors. 

To ensure the shape's naturalness, we use the B-spline curve \cite{cox1972numerical} to connect $i$-th patch's each vertex to generate the patch contour $M^{con}_i$ represented by:
\begin{equation}
    M^{con}_i = \{Bspline(p_{ij-1},p_{ij},p_{ij+1},p_{ij+2}|0 \leq j < n\}
    \label{Mcon}
\end{equation}

With $M^{con}_i$ being closed, we can easily fill $M^{con}_i$ and get the final $i$-th patch's shape $\mathcal{P}_i$. 

\textbf{Optimization Algorithm.}
To launch the attack under black-box settings, we employ the Differential Evolution (DE) algorithm to optimize the patch, which efficiently explores the search space to find the optimum for \cref{eq:untarget} and \cref{eq:target}. DE algorithm consists of four steps: starting from an initial population, using the crossover and mutation to generate the offspring population, making the fittest survive according to the fitness function, and finding the appropriate solution in the iterative evolution process.

In our case, a population represents a multi-patch attack strategy $\mathcal{P}_s=\{\mathcal{P}_1,\mathcal{P}_2,\dots,\mathcal{P}_m\}$. Given the population size $Q$, the $k$-th generation solutions $S(k)$ are represented as:
\begin{equation}
    S(k)=\{S_{i}(k)|\theta^{L}_{j}\leq S_{ij}(k)\leq \theta_{j}^{U}, 1\leq i\leq Q, 1\leq j\leq n\}
    \label{S(k)}
\end{equation}
where $S_i(k)$ is the shapes and positions of $i$-th multi-patch and $S_{ij}(k)$ is the parameters of shapes (\eg, $\mathcal{L}_s$) and positions (\eg, $\mathcal{O}_s$). Here $\theta^{L}_{j}$ and $\theta^{U}_{j}$ together make up the multi-feasible regions $\mathcal{M}_i$, which are the allowable range of the shape and location parameters. 
 


In the $k+1$ generation of DE, the solution $S(k +1)$ is achieved via crossover, mutation, and selection based on $S(k)$. A fitness function is applied on $S_i(k)$ to evaluate its attack effectiveness. During this process, the fitness function solely relies on the confidence scores of each identity. Consequently, black-box attacks can be performed to find the optimal multi-patch attack solution $\mathcal{P}_s^*$. The shape evolution process is shown at \cref{fig:shape}.

\setlength{\textfloatsep}{5pt}
\begin{algorithm}[t]  
    \caption{\small Shape Optimization for NIR Attack} 
    \label{alg:algorithm1}
        \KwIn{\small{Network $f$,  clean NIR image $x^{nir}$ and label $t$,  the fitness function $\mathcal{J}(\cdot)$, the objective function $\mathcal{L}(\cdot)$, population size $P$,  max iteration number $T$, face region mask $M^F$, value ranges $[l_{min},l_{max}]$}}
        \KwOut{\small{Adversarial NIR image $x_{adv}^{nir}$}}

        \small Initialize $S(0)$ randomly in deformation and position constraints, $\mathcal{J}(\cdot)=\mathcal{L(\cdot)}$
        
        \For{$k = 0$ to $T-1$}{
        
            \small Sort ${S}(k)$ in descending order based on $\mathcal{J}(S(k))$ 
            
            \If{${S_0(k)}$ makes the attack successful}{
            
            \small $stop=k$; break; 
            
            }
            
            \small Generate $S(k+1)$ based on crossover and  mutation

            \For{$i=1$ to $P$}{
            
            \small {Evaluate $S_{i}(k)$, $S_{i}(k+1)$ according to $\mathcal{J}(\cdot)$}            
            \small{$S_{i}(k+1)\leftarrow$ the better one in $S_{i}(k)$, $S_{i}(k+1)$} 
            
            }  
        
        }

        \small{Sort ${S}(stop)$ in descending order according to $\mathcal{J}(\cdot)$}

        \small{\textit{Pos} \& \textit{Def} parameters for $\mathcal{P}_s^*\leftarrow S_{0}(stop)$}
        
        \small{Calculate the each patchs' vertexs by \cref{caculate vertexs}}
        
        \small{$M^{con}\leftarrow$ combine each patch's vertexs by \cref{Mcon}}
        
        \small{$\mathcal{P}_s^*\leftarrow$ fill the $M_{con}$ 
        
        \small{$\mathcal{M}\leftarrow$ restricts the perturbation region $M^F \cap \mathcal{P}_s^*$ }
        
        \small{$x^{nir}_{adv}\leftarrow$ get final multi-patch by light reflection model}

        \small{\small\textbf{return} $x^{nir}_{adv}$}
}\end{algorithm}

\subsection{Ink Shaping Attack in the Physical Domain}
\label{sec:Ink shaping attack physical}
To generate robust adversarial examples in the physical domain, we implement the following method to minimize the gap between the digital domain and the physical domain:

\textbf{Light Reflection Model.} 
NIR images are formed by reflecting light from NIR LEDs off the skin, which makes the pixel values received by the NIR camera different from what we optimize in the digital world. To reduce this discrepancy, we model the light reflection equation as a BRDF function~\cite{schaepman2006reflectance}, which is represented by:
\begin{equation}
  f(\boldsymbol{l}, \boldsymbol{v}) = D_d  + \frac{D\left(\theta_{h}\right) F\left(\theta_{d}\right) G\left(\theta_{l}, \theta_{v}\right)}{4 \cos \theta_{l} \cos \theta_{v}}
\end{equation}
where $D_d$ is a diffuse component, $D(\theta_h)$ is Beckmann Microfacet Distribution Function describing the distribution of microfacets on a surface, $F(\theta_d)$ is Fresnel Reflection equation approximating how much light is reflected at different angles and $G(\theta_t,\theta_v)$ is Beckmann Geometry term accounting for the occlusion and shadowing effects that can occur during reflection. Specifically, given a NIR light intensity $I$ and the original NIR image $x^{nir}_{orig}$, the adversarial NIR image $x^{nir}_{adv}$ can be formulated by:
\begin{equation}
x^{nir}_{adv} = I\cdot f(\boldsymbol{l},\boldsymbol{v})\cdot x^{nir}_{orig}
\label{eq:X_adv}
\end{equation}

The overall algorithm for generating adversarial patches is summarized in \cref{alg:algorithm1}, and the whole illustration for this framework is given in \cref{fig: attack framework}.




\


\begin{table*}[htbp]
\setlength{\tabcolsep}{4.8pt}
    \centering
    \footnotesize
{

\caption{The results of attacks on the face recognition task. We report the attack success rate (ASR) of adversarial examples generated by our method and AiD~\cite{cohen2023accessorize} on the CASIA~\cite{li2013casia}, BUAA~\cite{huang2012buaa}, and Oulu-CASIA~\cite{zhao2011facial} datasets, evaluated against ResNeSt~\cite{zhang2022resnest}, LightCNN~\cite{wu2018lightCNN}, LightCNN-DVG~\cite{fu2021dvg}, and LightCNN-Rob~\cite{wu2019defending}. The gray cells represent white-box attack results, while the white cells represent black-box attack results. The optimal black-box attack results are highlighted in bold.} 
\label{tab: digital Performance comparisons}
\vspace{0.1in}
\small
\begin{tabular}{c c| c c c c| c c c c }
\hline
Method & \diagbox{Soure models}{Target models} & ResNeSt & LightCNN & DVG & Rob & ResNeSt & LightCNN & DVG & Rob \\
\hline

\multicolumn{2}{c|}{CASIA} &\multicolumn{4}{c|}{Dodging} &\multicolumn{4}{c}{Impersonation} \\ \hline

AiD& ResNeSt  &\cellcolor{gray!25} 97.64\% &12.46\% &0.00\% &0.00\% &\cellcolor{gray!25} 27.87\% &1.71\%&0.00\%&0.00\% \\

AiD & LightCNN  &39.35\%  &\cellcolor{gray!25}98.86\% &82.91\% &15.13\% &1.57\% & \cellcolor{gray!25}26.19\% &17.81\%&0.28\% \\

AiD & DVG  &37.96\% &94.82\% & \cellcolor{gray!25}96.21\%&22.13\% &1.42\% &20.06\%& \cellcolor{gray!25}25.07\% &2.54\% \\

 AiD& Rob  &36.55\% &66.11\% &56.16\% &\cellcolor{gray!25} 61.49\% & 1.26\%&7.54\%&10.66\%& \cellcolor{gray!25}14.29\% \\ \arrayrulecolor{lightgray} \hdashline 

\arrayrulecolor{black}
Ours &  & \textbf{89.29}\% & \textbf{97.44}\% &\textbf{84.85}\% &\textbf{62.91}\% &\textbf{76.96}\%&\textbf{66.85}\% &\textbf{33.01}\% &\textbf{35.11}\% \\

\hline

\multicolumn{2}{c|}{BUAA} &\multicolumn{4}{c|}{Dodging} &\multicolumn{4}{c}{Impersonation} \\ \hline

AiD & ResNeSt & \cellcolor{gray!25}99.26\% &0.00\% &0.00\% &0.00\% &\cellcolor{gray!25}100.00\% & 0.00\% & 0.00\% & 0.00\%  \\

AiD & LightCNN &26.53\% &\cellcolor{gray!25}99.32 \% & 82.43\%& 25.85\% &5.14\% &\cellcolor{gray!25}98.64\% &\textbf{82.31}\% &14.48\% \\

AiD & DVG &25.17\% &\textbf{96.59}\% & \cellcolor{gray!25}98.64\% &38.09\% &4.41\% &\textbf{87.76}\% &\cellcolor{gray!25} 95.24\% &20.69\% \\

AiD & Rob &7.43\% &27.21\% &39.46\% &\cellcolor{gray!25}80.69\% &3.68\% &19.04\% &30.61\%  & \cellcolor{gray!25}78.23\%  \\ \arrayrulecolor{lightgray} \hdashline 
\arrayrulecolor{black}
Ours & &\textbf{64.71}\% &91.16\% &\textbf{83.67}\% & \textbf{42.76}\% &\textbf{44.89}\% &40.14\% &38.09\% &\textbf{32.43}\%\\

\hline
\multicolumn{2}{c|}{Oulu-CASIA} &\multicolumn{4}{c|}{Dodging} &\multicolumn{4}{c}{Impersonation} \\ \hline

AiD & ResNeSt &\cellcolor{gray!25} 89.33\% &17.94\% &20.51\% &29.49\% & \cellcolor{gray!25}90.82\%&10.67\%&11.54\%&11.84\% \\

AiD & LightCNN &26.92\% & \cellcolor{gray!25}98.73\% &84.62\% &41.03\% &13.33\% &\cellcolor{gray!25}91.03\%&\textbf{79.49}\%&7.89\% \\

AiD & DVG&17.95\% &89.74\% &\cellcolor{gray!25} 91.03\% &48.72\% &10.67\% &\textbf{87.18}\% &\cellcolor{gray!25} 92.31\% &22.37\%  \\

AiD & Rob &19.74\% &61.84\% &63.16\% &\cellcolor{gray!25}77.63\% &8.00\% &42.31\% &56.41\% &\cellcolor{gray!25}69.74\% \\ \arrayrulecolor{lightgray} \hdashline 
\arrayrulecolor{black}

Ours&  &\textbf{72.00}\% &\textbf{92.31}\% &\textbf{89.74}\% &\textbf{78.95}\%&\textbf{58.67}\%& 67.94\% & 65.38\% &\textbf{61.84}\% \\
\hline
\end{tabular}}

\end{table*}

\section{Experiments}
This section begins by presenting the experimental setup and then compares our multi-patch adversarial attack with the state-of-the-art (SOTA) method proposed by Cohen and Sharif~\cite{cohen2023accessorize} in both the digital and physical domains.

\subsection{Experiment Setup}
\label{sec: Experiment Setup}
\noindent \textbf{Datasets:} 
We conduct evaluations on three benchmark NIR-VIS datasets: the CASIA NIR-VIS 2.0 face database \cite{li2013casia}, the BUAA-VisNir face database \cite{huang2012buaa}, and the Oulu-CASIA NIR-VIS facial expression database \cite{zhao2011facial}. The \textbf{CASIA NIR-VIS 2.0} with 725 subjects (1-22 VIS/5-50 NIR images per subject) captured under visible light and 850nm NIR illumination, exhibiting pose, expression, and distance variations. The \textbf{BUAA-VisNir} contains 150 subjects with synchronized VIS-NIR pairs via multispectral imaging under controlled illumination gradients. And the \textbf{Oulu-CASIA} (\cite{zhao2011facial}) features 80 subjects performing six expressions under three illuminations with 25fps video sequences. 

\noindent \textbf{Evaluation method:} We evaluate the proposed method by attacking the identification results from state-of-the-art NIR face recognition models, which includes ResNeSt~\cite{zhang2022resnest}, LightCNN~\cite{wu2018lightCNN}, LightCNN-DVG~\cite{fu2021dvg}, and LightCNN-Rob~\cite{wu2019defending}, all achieving 100\% baseline identification accuracy under benign conditions. Attack Success Rate (ASR) is quantified through two critical metrics:

\begin{itemize}
    \item Dodging: Aims to evaluate the method's ability to evade correct recognition by suppressing the confidence score of the ground-truth identity, ensuring it is not selected as the system’s top-1 prediction.
    \item Impersonation: Aims to evaluate the method's ability to force target recognition by increasing the target identity's confidence score relative to others, ensuring it becomes the top-1 prediction.
\end{itemize}

\begin{figure}[t]
\centering
\input{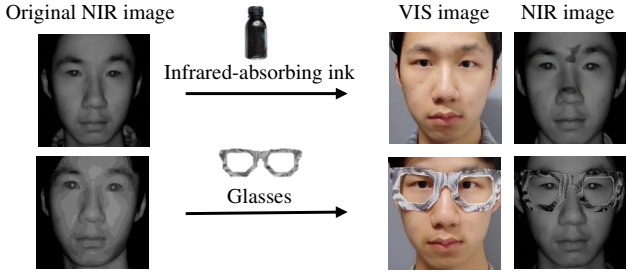}

\caption{Experiment preparation for our method and AiD~\cite{cohen2023accessorize}}

\label{fig:attack comparison}
\end{figure}

We also benchmark against Accessorize in the Dark (AiD)~\cite{cohen2023accessorize}, the current state-of-the-art (SOTA) physical attack on NIR face recognition systems, by comparing ASRs in both digital and physical domains, with experiments conducted by volunteers, as shown in \cref{fig:attack comparison}.

%

\noindent \textbf{Implementation:} To determine the patch position, we use the \texttt{dlib} library to extract 81 facial feature points and fill the effective region to generate the mask $M^F$. The shape parameters for the attack are set as follows: the number of patches $m$ is 4, the number of vertex $n$ is 8, and the value range of shape deformation constraint $[l_{\min}, l_{\max}] = [2, 20]$. The optimization algorithm parameters are configured with a population size $P=40$, and the maximum number of iterations $T=200$.



\subsection{Digital Domain Attacks}
\label{sec: digital domain attacks}

\begin{figure}[t]
\centering

\input{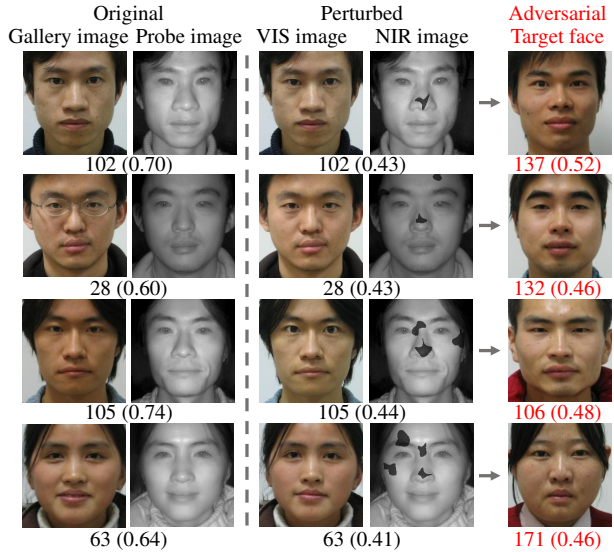}

\caption{Examples of digital attacks using ink-shaping attack. For each group, the five images correspond to the gallery VIS image, the unattacked original image, the VIS and NIR image after the attacks, and the VIS image corresponding to the predicted wrong class after the attacks. Black text denotes the original correct ID and its probability, while red text indicates the misclassified ID and its confidence after the attack.}

\label{fig: digit attack images}
\end{figure}

We first perform white-box and transfer-based black-box attacks using the AiD approach and compare their attack success rate (ASR) with our method in the black-box setting, which only relies on the models' predicted labels and probabilities.
The results are presented in \cref{tab: digital Performance comparisons}. We first compare our black-box attacks with AiD's transfer-based black-box attacks. In the black-box setting, our method attains an average ASR of 65.45\% and 51.77\% performing dodging and impersonation attacks, respectively, outperforming AiD's average ASR of 37.61\% and 19.30\%. Although constrained by limited information, our method's average ASR is only reduced by 13.64\% compared to the white-box AiD attacks, which have complete knowledge of the model's structure and parameters. Additionally, four groups of visual examples are shown in \cref{fig: digit attack images}. As observed, our method effectively decreases the probability of the ground-truth identity, ensuring that the probabilities of others exceed it, thus achieving a successful attack with minimal attention. These findings highlight that our method achieves more effective attacks across various network architectures compared to the SOTA method (AiD) in the digital domain.


\begin{figure*}[t]
\centering
\input{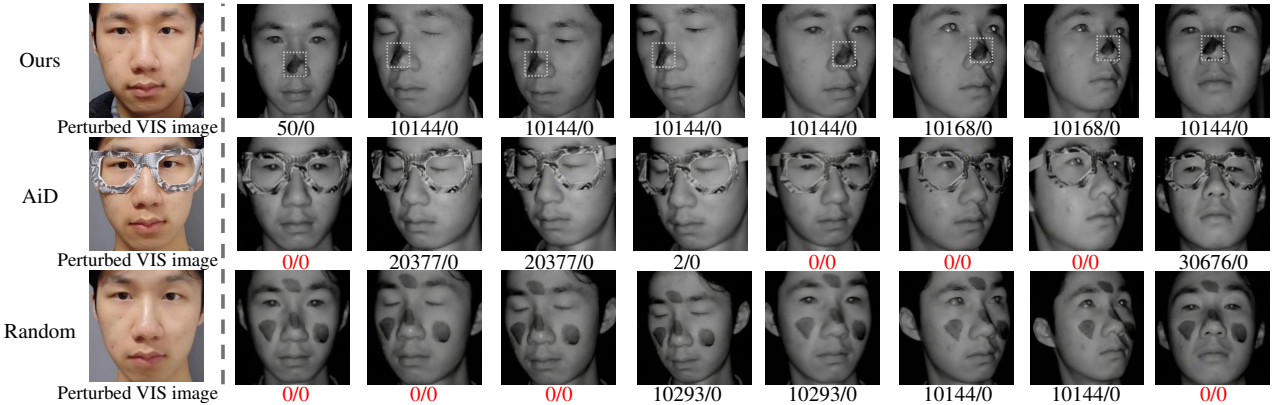}

\caption{The examples demonstrate physical attacks on NIR face recognition models. They present VIS images perturbed with infrared-absorbing ink or glasses and evaluate attack effectiveness across different facial postures using our method, AiD~\cite{cohen2023accessorize}, and random method. The text below each image displays the predicted and ground-truth labels, with failed attacks highlighted in red. The perturbed ink areas are highlighted by white rectangular boxes. }
\label{fig: physical attack}

\end{figure*}

\subsection{Physical Domain attacks}
\label{sec: physical domain attacks}
Here we utilize three methods to implement physical attacks: ours, AiD, and random. We first optimize the patches's parameters using the front-facing NIR image in the digital domain by our method and AiD, then perform attacks in the physical domain, as shown in \cref{fig:attack comparison}. Additionally, random method attacks are triggered by the stochastic application of infrared-absorbing ink to various areas of the facial regions. Following \cite{wei2022adversarial}, we recorded a 35-second video at 10 frames per second (approximately 350 frames in total), capturing NIR images with adversarial perturbations under varying face postures (e.g., front-facing, upward gaze, and rotations within a 30-degree range), as shown in \cref{fig: physical attack}. As observed, our approach achieved successful attacks across various face postures while introducing minor perturbations in the perturbed NIR images and remaining human-imperceptible in the perturbed VIS images. The quantitative results are shown in \cref{tab: Physical attack report}. The average ASR of our method is 82.46\%, significantly outperforming the AiD method (64.18\%) and the random method (45.84\%). This superiority stems from optimizing shape and position, which, unlike color optimization, are more robust in physical attack scenarios as they remain relatively stable in real-world conditions. Moreover, the results also confirm that only optimized shapes and positions can exploit model vulnerabilities and effectively execute physical attacks. Additionally, we compute the percentage of frames under different face postures, as shown in \cref{fig: physical attack different posutre} (a). Our method demonstrates superior robustness in maintaining attack effectiveness across different face postures, outperforming both AiD and the random method, except under extreme deformations or near-invisibility. This highlights the effectiveness of our attack framework in addressing the optimization mismatch between digital and real-world domains, ultimately leading the patch toward its physically optimal shape and position. Furthermore, we also conduct dodging and impersonation attacks in the physical domain. As shown in \cref{tab: physical dodging and impersonation report}, the ground-truth labels' probabilities in different models are significantly reduced, which proves that the generated attack parameters in the digital domain can still maintain a good attack performance when applied to the physical domain.

\begin{table}[t]
\centering
    \caption{The percentage of video frames successfully attacked when continuously changing face postures in the physical domain.}
    \vspace{0.1in}
   \small
\begin{tabular}{c c c c c  }
  \hline  
Models        &  ResNeSt & LightCNN & DVG & Rob \\   \hline  
Ours  & 95.42\% &83.49\% & 80.05\% &70.82\%  \\    
AiD   & 100.00\%&55.44\% &22.54\% & 78.74\%\\ 
Random & 90.91\% & 28.86\% & 13.89\% &50.28\%\\ \hline
\end{tabular} 
\label{tab: Physical attack report}

\end{table}

\begin{table}[t]
\centering
   \caption{The probabilities of the ground-truth label 0 before and after the attack in the physical domain.}
\vspace{0.1in}
   \small
\begin{tabular}{c c c c c}
  \hline  
Models    &  ResNeSt & LightCNN & DVG & Rob  \\ \hline
Attack type &\multicolumn{4}{c}{Dodging} \\ \hline

    Before-attack & 1.00 &0.59 &0.74 &0.55 \\
    After-attack & 0.32 &0.38 &0.36 & 0.36\\
    Difference($\downarrow$) & 0.68 &0.11&0.38&0.19 \\ 
    \hline
Attack type& \multicolumn{4}{c}{Impersonation} \\ \hline
    Before-attack & 1.00 & 0.99 & 1.00&0.55\\
    After-attack & 0.26 & 0.27 & 0.33 &0.23\\
    Difference($\downarrow$) & 0.74&0.72&0.67&0.22 \\ 
    \hline
\end{tabular} 

\label{tab: physical dodging and impersonation report}
\end{table}

\section{Ablation Study}
This section presents ablation studies to assess the impact of various factors on optimizing adversarial patch attacks, including position and shape optimization and the Light Reflection Model strategy.

\subsection{Effects of Position and Shape Optimization}
\label{sec: Effects of Position and Shape Optimization}
To investigate the effects of our optimized positional attributes, we randomly select several positions for subsequent shape-only optimization; To examine the influence of shape optimization, we first generate adversarial multi-patches to determine the optimal positions for a given sample and then alter the optimal shapes to basic forms such as circles, squares, rectangles, and triangles. The results are presented in \cref{tab: position and shape comparision}. The combined shape and position optimization method achieves an average ASR of 83.25\%, outperforming the shape-only optimization by 66.07\% and the position-only optimization by 63.34\%.  These findings highlight the critical role of shape and position attributes in enhancing the effectiveness of our adversarial attacks.
\begin{table}[h]

\small
\centering
   \caption{The ASR comparison with different optimization attributes for dodging attack across various models.}
 \vspace{0.1in}
\begin{tabular}{c c c c}
  \hline  
Opt. attributes    &  Shape + Position & Shape & Position \\   \hline  
ResNeSt     &       \textbf{ 72.00}\%   &  4.00\% &18.39\%      \\    
LightCNN &  \textbf{92.31}\%   &  17.95\% &14.36\% \\ 
DVG   & \textbf{89.74}\% &  23.08\% &12.05\% \\ 
Rob   & \textbf{78.95}\%  &  23.68\% &34.87\% \\ \hline
\end{tabular} 

\label{tab: position and shape comparision}
\end{table}

\subsection{Effects of Light Reflection Model Strategy}
\label{sec: Effects of Light Reflection Imaging Model Strategy}

\begin{figure}[t]
\centering
\input{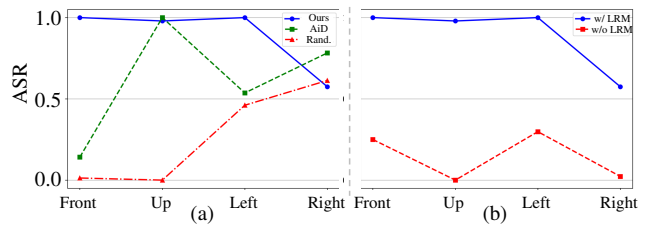}

\caption{The comparison of attack success rates across different face postures. Subfigure (a) compares three physical attack methods: Ours, AiD, and Random. Subfigure (b) compares the optimal results of our approach under two conditions: with and without LRM.}

\label{fig: physical attack different posutre}
\end{figure}

Considering the imaging differences between the digital and physical domains, we propose a Light Reflection Model (LRM) to generate more effective shapes. To evaluate the effectiveness of the LRM, we optimized our patch both with and without the LRM. When using the LRM, we followed the method proposed in \cref{sec:Ink shaping attack physical} to generate the adversarial NIR image. In the absence of the LRM, we simply set the pixel values of the patch region to zero. The quantitative results, presented in \cref{tab:LRM comparison}, indicate a 37.61\% increase in the average ASR with LRM strategy, compared to its absence. Furthermore, we calculate the percentage of successful attack frames across various face postures in the physical domain, as shown in \cref{fig: physical attack different posutre} (b). The experimental validation confirms the crucial role of LRM in achieving the posture robustness of our patch, since without LRM optimization, the ASR dropped sharply by more than 90\% in both upward and rightward poses. These results highlight the critical role of LRM in achieving physically optimal adversarial patch attacks.

\begin{table}[t]
\centering

   \caption{The percentage of video frames successfully attacked optimization with or without LRM strategy in the physical domain.}
   \vspace{0.1in}
   \small
\begin{tabular}{c c c | c}
  \hline  
Optimization methods        &  w/ LRM & w/o LRM & Difference  \\   \hline  
ResNeSt    &94.86\%& 84.38\% & $\downarrow$10.48\%    \\    
LightCNN    &83.49\%& 18.50\% & $\downarrow$64.99\%    \\    
DVG         &80.05\%& 26.37\% & $\downarrow$53.68\%    \\    
Rob & 70.82\% & 49.54\% & $\downarrow$21.28\% \\ \hline
\end{tabular} 

\label{tab:LRM comparison}
\end{table}

\section{Limitation}
In this study, we focused on adversarial patch attacks targeting NIR face recognition systems equipped with 850 nm wavelength filters, as they are widely used. However, the presence of NIR cameras operating at different wavelengths (\eg, 810 nm, 850 nm) can lead to variations in imaging characteristics, potentially affecting our patch's performance. In future work, we will explore the development of more generalized adversarial patches capable of operating across a broader range of wavelengths. This would involve testing and optimizing patches for various NIR wavelength filters, including 810 nm, 850 nm, and others, to enhance the robustness of attacks across diverse camera setups.

\section{Conclusion}
In this paper, we propose a novel human-imperceptible physical adversarial attack against NIR face recognition models, emphasizing the security risks of these systems in real-world deployments. By leveraging infrared-absorbing ink, we design a stealthy adversarial patch that disrupts NIR imaging while remaining visually inconspicuous. Our method optimizes both the shape and position of adversarial patches using a differential evolution algorithm and further enhances transferability from digital to physical domains through a light reflection model. Experimental results demonstrate that our approach achieves significantly higher ASR than the SOTA method. These findings highlight the vulnerability of NIR face recognition systems and underscore the need for more robust defense mechanisms. 


{\small
\bibliographystyle{ieee_fullname}
\bibliography{egpaper_for_review}
}

\end{document}